\documentclass[10pt,conference]{IEEEtran}
\IEEEoverridecommandlockouts
\usepackage{cite}
\usepackage{amsmath,amssymb,amsfonts}
\usepackage{algorithm}
\usepackage{graphicx}
\usepackage{textcomp}
\usepackage{xcolor}
\usepackage{url}
\usepackage{hyperref}
\usepackage{subfig}
\usepackage{tabularx}
\usepackage{algpseudocode}
\usepackage{setspace}
\usepackage{etoolbox}
\usepackage{booktabs}
\usepackage{stfloats}
\usepackage{caption}
\usepackage{listings}
\usepackage{changepage} 
\usepackage{xcolor}
\usepackage{placeins} 
\usepackage{float}

\usepackage{tcolorbox}
\tcbset{
  colback=gray!5,
  colframe=gray!50,
  boxrule=0.3pt,
  left=3pt,
  right=3pt,
  top=1pt,
  bottom=1pt,
  width=0.96\linewidth,
  sharp corners,
  before skip=6pt,   
  after skip=6pt     
}

\usepackage{inconsolata}
\newcommand{\code}[1]{\texttt{\small #1}}

\algrenewcommand\algorithmicindent{1.5em} 

\begin{document}

\title{Neuro-Symbolic Compliance: Integrating LLMs and SMT Solvers for Automated Financial Legal Analysis}

\author{
Yung-Shen Hsia\IEEEauthorrefmark{1},
Fang Yu\IEEEauthorrefmark{1},
and Jie-Hong Roland Jiang\IEEEauthorrefmark{2} \\[4pt]
\IEEEauthorblockA{\IEEEauthorrefmark{1}Department of Management Information Systems, National ChengChi University, Taipei, Taiwan\\
Emails: 113356046@g.nccu.edu.tw, yuf@nccu.edu.tw}
\\[4pt]
\IEEEauthorblockA{\IEEEauthorrefmark{2}Department of Electrical Engineering, National Taiwan University, Taipei, Taiwan\\
Email: jhjiang@ntu.edu.tw}
}

\maketitle

\begin{abstract}
Financial regulations are increasingly complex, hindering automated compliance—especially the maintenance of logical consistency with minimal human oversight. We introduce a \textbf{Neuro-Symbolic Compliance Framework} that integrates Large Language Models (LLMs) with Satisfiability Modulo Theories (SMT) solvers to enable \textit{formal verifiability} and \textit{optimization-based compliance correction}. The LLM interprets statutes and enforcement cases to generate SMT constraints, while the solver enforces consistency and computes the \textit{minimal factual modification} required to restore legality when penalties arise. Unlike transparency-oriented methods, our approach emphasizes \textbf{logic-driven optimization}, delivering verifiable, legally consistent reasoning rather than post-hoc explanation. Evaluated on 87 enforcement cases from Taiwan’s Financial Supervisory Commission (FSC), the system attains 86.2\% correctness in SMT code generation, improves reasoning efficiency by over 100×, and consistently corrects violations—establishing a preliminary foundation for optimization-based compliance applications.
\end{abstract}

\begin{IEEEkeywords}
Financial precedent analysis, Retrieval-Augmented Generation (RAG), LLM Agent, SMT Solver, Neuro-Symbolic, Legal tech, Legal understanding analysis, Constraint solving, Financial Supervisory Commission (FSC),Optimization, Legal consistency, MaxSMT
\end{IEEEkeywords}

\begin{figure*}[!t]
  \centering
  \includegraphics[width=\linewidth]{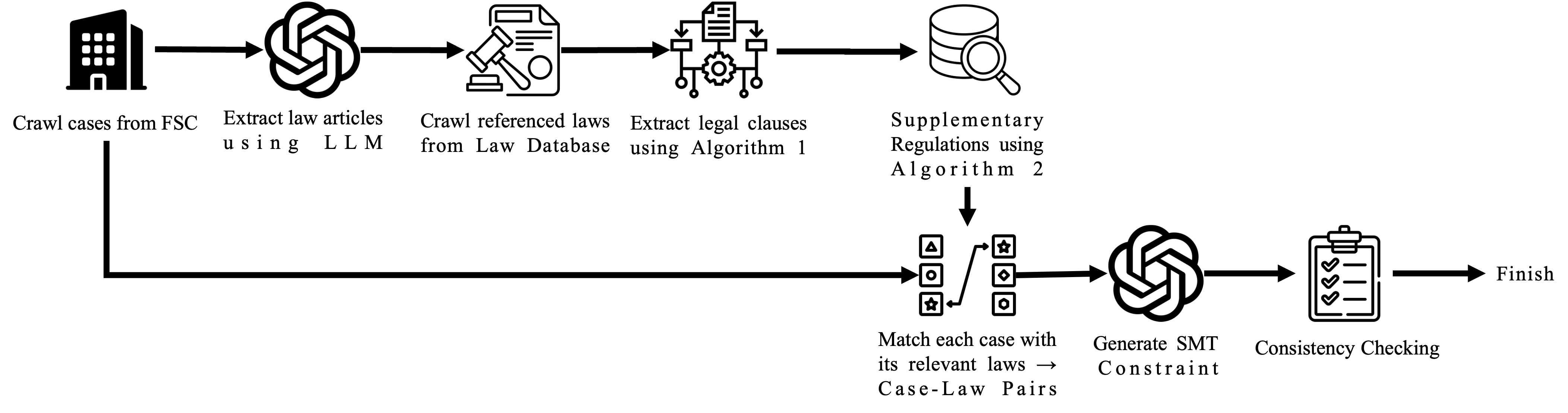}
  \caption{SMT Constraint Generation Pipeline}
  \label{fig:flowchart}
\end{figure*}

\section{Introduction}

As legal regulations grow in complexity and change more frequently, automating compliance has become a key challenge in Legal Tech~\cite{TowardstransparentlegalFormalization,ponkin2019digital}. A central difficulty lies in translating legal provisions into machine-readable formats while preserving their semantic richness~\cite{surden2024computable}. Meanwhile, AI has accelerated transformation in finance, supporting tasks from anomaly detection to regulatory monitoring~\cite{mohsen2024digital,bolton2002statistical,arner2015evolution}.

Formal methods like SAT/SMT solving and model checking~\cite{barrett2011cvc4,de2008z3} have recently been applied to legal reasoning by encoding laws and facts as constraints~\cite{bradley2013presidential}. Large Language Models (LLMs) perform well in interpreting legal text~\cite{athul2024legalmind,shu2024lawllm} but often lack precision for high-stakes tasks. This motivates hybrid neuro-symbolic approaches that combine the interpretability of symbolic reasoning with the language capabilities of LLMs~\cite{hitzler2022neuro,mao2025neuro}.

Neuro-symbolic AI enables compositional generalization and transparency by integrating logic-based reasoning with deep learning. It has shown success in areas such as visual reasoning, robotic manipulation, and concept grounding~\cite{mao2025neuro}. These capabilities are especially relevant in compliance domains, where decisions must be both explainable and legally defensible.

Recent efforts in combining LLMs with formal reasoning systems have underscored the growing demand for verifiable AI reasoning. 
For instance, the LLM-Based Formalized Programming (LLMFP) framework~\cite{hao2024planning} demonstrates that translating natural language problems into Satisfiability Modulo Theories (SMT) formulations enables completeness and soundness guarantees that purely neural approaches cannot ensure. Similarly, Chen et al.~\cite{chen2024can} show that even advanced prompting strategies, such as Dual Chains of Logic (DCoL), still fall short of achieving consistent logical inference, as LLMs inherently lack self-verification mechanisms. These findings suggest that while LLMs excel in flexible natural language understanding, symbolic solvers provide the rigor required for provable correctness. 

Building on this insight, Our work integrates LLMs with SMT-based formal verification to bridge interpretability and verifiability—aiming to enhance both transparency and logical consistency in regulatory compliance tasks.

Complementing this trend, the study on \textit{Formal Verification of LLM-based Agents}~\cite{zhang2024fusion} further emphasizes that LLM reasoning processes should be formally checkable, and proposes strategies such as \textit{Multiple LLMs Debating} and \textit{Self-Correction} to enhance consistency and reliability. 
This collective direction highlights a broader consensus that coupling LLMs with formal methods is essential for developing trustworthy and verifiable AI systems.

Yet, few frameworks target the unique characteristics of financial legal enforcement in specific jurisdictions. Taiwan’s regulatory ecosystem, for example, involves multilingual documentation, localized statutes, and detailed enforcement case records. These pose challenges that are not well addressed by existing compliance tools.

In this work, we present a compliance framework that combines LLMs with SMT-based verification, tailored to Taiwan’s financial legal context. A RAG pipeline integrates statutes, enforcement cases, and violation records from the FSC. Legal norms are encoded as constraints, and generative models aid interpretation to support violation prediction and legal guidance.

We evaluate the system on 87 real-world FSC enforcement cases, demonstrating its ability to detect unlawful contract terms and generate rule-consistent recommendations. The SMT layer enhances interpretability and accuracy, offering a potential foundation for more transparent and verifiable Legal Tech applications.

\section{Related Work}
Numerous studies have explored the automation of regulatory compliance through various computational techniques. LLMs have gained attention in this space. In financial auditing, Berger et al.~\cite{10386518} evaluated LLMs for automated compliance checks, comparing models such as LLaMA-2 and GPT-4. GPT-4 excelled in multilingual tasks, while LLaMA-2 was more cost-effective for detecting non-compliance. However, challenges like variability and false positives persist. Similarly, Liao et al.~\cite{liao2025neural} propose a Neuro-Symbolic Reasoning framework, where LLMs generate logical forms that are verified by symbolic solvers. Ning et al.~\cite{ning2025gns} expand on this by integrating multi-agent planning and external symbolic tools for enhanced reasoning and explainability. Shi et al.~\cite{10888935} also present a dual-system neuro-symbolic architecture, combining heuristic LLM inference with logic-based verification to improve accuracy and interpretability. These efforts show growing convergence between language models and formal methods for regulatory tasks.

SMT solvers and symbolic approaches offer another line of research. Judson et al.~\cite{judson2024soid} introduced \textit{soid}, which integrates Z3 with symbolic execution via KLEE, providing legal practitioners with GUI-supported formal accountability tools. Feng et al.~\cite{feng2023early} translated legal constraints into SMT using Metric First-Order Temporal Logic (MFOTL), enabling efficient compliance checks in healthcare and finance. Similarly, Tsigkanos et al.~\cite{tsigkanos2019firms} applied SMT to model economic networks, ensuring consistency through logical constraints.

Hybrid LLM--SMT systems aim to bridge natural and formal language. The Semantic Self-Verification (SSV) framework~\cite{razaalmost} combines LLM-generated constraints with SMT solver feedback, improving accuracy on formal reasoning tasks. This integration minimizes manual effort while enhancing correctness. Other complementary methods support this landscape. Beach et al.~\cite{beach2015rule} proposed a semantic framework that decouples legal expertise from implementation, empowering non-technical users in compliance modeling. Separately, counterfactual explanations~\cite{wachter2017counterfactual,andreas2023counterfactuals} have been studied for making algorithmic decisions transparent and contestable under data protection laws such as GDPR.

Our work builds upon these trends by applying LLM--SMT integration to regulatory compliance. Rather than focusing solely on translation or dialogue modeling, we construct an end-to-end pipeline that supports constraint synthesis, verification, and interactive correction, offering interpretable and verifiable legal analysis.

\section{Methodology}

\subsection{SMT Constraint Generation}

We first discuss our SMT constraint generation pipeline. Figure~\ref{fig:flowchart} illustrates the overall process of transforming unstructured legal and case documents into formalized SMT constraints. The pipeline begins with data collection and preprocessing, where legal articles and case texts are parsed and structured. Subsequently, key legal elements and logical relationships are extracted to construct high-quality constraints suitable for automated reasoning.

\subsubsection{Case Crawling and Preliminary Article Extraction}

We collected financial regulatory case summaries from public repositories, each containing structured metadata and a narrative describing the violation.  
We use an LLM to extract the legal provisions explicitly mentioned in the case description.  

\subsubsection{Legal Article Retrieval and Parsing}

For each cited provision, the corresponding legal text was retrieved from an official law database and parsed using a rule-based algorithm (Algorithm~\ref{alg:extract_articles}). The extracted articles were structured into dictionaries mapping article numbers to their titles, clauses, and full content.

Each case was then mapped to its cited articles, forming a structured case–law association dataset suitable for semantic retrieval and reasoning.

\begin{algorithm}[ht]
\caption{Article and Clause Extraction from Raw Legal Documents}
\label{alg:extract_articles}
\begin{algorithmic}[1]
\Require Raw document $D$
\Ensure Dictionary $A$ mapping article numbers to titles, clauses, and content
\State Initialize empty text $T \gets ""$
\State Open $D$ using \code{pdfplumber}; extract and concatenate text into $T$
\State Split $T$ into lines $L$
\State Initialize empty dictionary $A$; set $a \gets \code{None}$, $c \gets \code{None}$
\For{each line $l$ in $L$}
    \State Normalize $l$ (strip and collapse whitespace)
    \If{$l$ matches section heading pattern}
        \State \textbf{continue}
    \EndIf
    \If{$l$ matches ``Article $n$''}
        \State $a \gets n$; initialize $A[a]$ with title, empty clauses, and content
        \State $c \gets \code{None}$
    \ElsIf{$a \neq \code{None}$}
        \If{$l$ matches clause pattern (e.g., ``$m.$ text'')}
            \State Append clause text to $A[a][\code{clauses}]$; set $c \gets text$
        \ElsIf{$c \neq \code{None}$}
            \State Append $l$ to last clause in $A[a][\code{clauses}]$
        \Else
            \State Append $l$ to $A[a][\code{content}]$
        \EndIf
    \EndIf
\EndFor
\Return $A$
\end{algorithmic}
\end{algorithm}

\subsubsection{Supplementary Article Retrieval (RAG-Based Enrichment)}
\label{sec:legal-rag-augmentation}

To ensure comprehensive coverage of legal context, a Retrieval-Augmented Generation (RAG) module expands each base article $L_0$ by retrieving semantically related provisions and enforcement rules from a legal database~$\mathcal{D}$. This procedure improves SMT constraint completeness by including contextually supportive clauses.

The process, detailed in Algorithm~\ref{alg:legal-rag-augmentation}, consists of four stages:

\begin{enumerate}
  \item \textbf{Query Generation:}  
  An LLM generates a diverse query set $\mathcal{Q}=\{q_1,\dots,q_m\}$ to capture different semantic aspects of $L_0$.
  \item \textbf{Hybrid Retrieval:}  
  Each query $q$ performs both BM25 and vector searches:
    \begin{equation*}
    \begin{split}
    R_{\mathrm{bm25}} &= \mathrm{BM25Search}(\mathcal{D}, q, K),\\
    R_{\mathrm{vec}}  &= \mathrm{VectorSearch}(\mathcal{D}, q, K).
    \end{split}
    \end{equation*}
    Results are merged as $R = R_{\mathrm{bm25}} \cup R_{\mathrm{vec}}$, and scored by a hybrid relevance function:
    \begin{equation*}
    s_{\mathrm{hybrid}}(d, q) 
    = \alpha \, \mathrm{sim}_{\mathrm{vec}}(d, q)
    + (1-\alpha) \, \mathrm{score}_{\mathrm{bm25}}(d, q),
    \end{equation*}
    where $\alpha = 0.8$ balances semantic and lexical relevance.
  \item \textbf{Reranking and Filtering:}  
  The top-$K$ candidates are reranked by a Cross-Encoder (\code{FlagReranker}) and filtered via \code{LLM\_FilterUseful} to retain non-redundant, contextually relevant clauses.
  \item \textbf{Aggregation:}  
  Filtered clauses $C$ from all queries are merged and deduplicated into a final supplementary set $\mathcal{S}$.
\end{enumerate}

This retrieval module, implemented as the \textit{LegalSearchEngine}, integrates semantic and lexical retrieval, hybrid scoring, reranking, and LLM filtering. It employs \textit{text-embedding-ada-002} for embeddings, \textit{ChromaDB} for vector indexing, and \textit{BAAI/bge-reranker-v2-m3} for reranking.

\begin{algorithm}[ht]
\caption{Supplementary Article Expansion via Retrieval-Augmented Generation}
\label{alg:legal-rag-augmentation}
\begin{algorithmic}[1]
\Require Base article $L_0$; database $\mathcal{D}$; weight $\alpha \in [0,1]$; retrieval size $K$
\Ensure Supplementary set $\mathcal{S}$
\State $\mathcal{Q} \gets \mathrm{LLM\_GenQueries}(L_0)$
\State $\mathcal{S} \gets \emptyset$
\For{each $q \in \mathcal{Q}$}
    \State $R_{\text{bm25}} \gets \mathrm{BM25Search}(\mathcal{D}, q, K)$
    \State $R_{\text{vec}} \gets \mathrm{VectorSearch}(\mathcal{D}, q, K)$
    \State $R \gets R_{\text{bm25}} \cup R_{\text{vec}}$
    \For{each $d \in R$}
        \State Compute hybrid score:
        \Statex \hspace{2em}$s_{\text{hybrid}}(d) =
          \alpha \cdot \mathrm{sim}_{\text{vec}}(d,q)
          + (1-\alpha) \cdot \mathrm{score}_{\text{bm25}}(d,q)$
    \EndFor
    \State $R_{\text{top}} \gets \mathrm{TopK}(R, s_{\text{hybrid}}, K)$
    \State $R' \gets \mathrm{Rerank}(R_{\text{top}}, \mathrm{CrossEncoder}, q)$
    \State $C \gets \mathrm{LLM\_FilterUseful}(R', L_0)$
    \State $\mathcal{S} \gets \mathcal{S} \cup C$
\EndFor
\Return $\mathrm{Unique}(\mathcal{S})$
\end{algorithmic}
\end{algorithm}

\subsubsection{SMT Constraint Generation}

After obtaining the case–article pairs, the system proceeds to generate SMT (Satisfiability Modulo Theories) constraints that formally represent the relationship between the factual scenario and the corresponding legal provisions.  
Once the constraints are generated, a series of validation and consistency checks are performed to ensure their correctness and logical soundness.  

\paragraph{Build Law Constraints}
Legal reasoning requires that all actions conform to statutory norms. To formalize these rules, we use LLM interpretation to convert legal provisions into Boolean expressions representing their logical structure.  
Each statutory clause is expressed as a set of propositional formulas—typically involving conjunctions, disjunctions, and implications—to capture the conditional relationships defined by the law.

To illustrate the process, we present a real case from the Financial Supervisory Commission (FSC).\footnote{\url{https://www.fsc.gov.tw/en/home.jsp?id=131&parentpath=0,2&mcustomize=multimessages_view.jsp&dataserno=202407120004&dtable=Penalty}}  
Using our extraction and RAG modules (Section~\ref{sec:legal-rag-augmentation}), relevant provisions were retrieved from the \textit{Insurance Act} (Articles~143-4 and~143-6)\footnote{\url{https://law.moj.gov.tw/ENG/LawClass/LawAll.aspx?pcode=G0390002}} and its \textit{Enforcement Rules} (Articles~2–5).\footnote{\url{https://law.moj.gov.tw/LawAll.aspx?pcode=G0390003}}  
These articles define the capital adequacy classification and corresponding supervisory measures. The encoded SMT hard constraints are summarized below.

Let \(r = \tfrac{\textit{own\_capital}}{\textit{risk\_capital}} \times 100\).  
Then the capital level classification is defined as:
\[
\textit{capital\_level} =
\begin{cases}
4, & r < 50 \text{ or } \textit{net\_worth}<0,\\
3, & 50 \le r < 150,\\
2, & 150 \le r < 200,\\
1, & r \ge 200.
\end{cases}
\]

Supervisory measures are expressed as:
\[
\begin{aligned}
c\textit{L2\_exec} &\Leftrightarrow
(\textit{plan\_submitted} \land \textit{plan\_executed}),\\
\textit{L3\_exec} &\Leftrightarrow
(\textit{L2\_exec} \land \textit{person\_removed} 
\land \textit{asset\_approved}).
\end{aligned}
\]

The penalty rule is defined compactly as:
\[
\begin{aligned}
\textit{penalty}=1 \text{ if } &
(\textit{level}=4 \land \neg\textit{L4\_exec})\\[-1pt]
&\vee\,(\textit{level}=3 \land \neg\textit{L3\_exec})\\[-1pt]
&\vee\,(\textit{level}=2 \land \neg\textit{L2\_exec}),\\[3pt]
\textit{penalty}&=0 \text{ otherwise.}
\end{aligned}
\]

The resulting set of formulas constitutes the law constraint base, which serves as the foundation for SMT encoding and subsequent logical verification.

\paragraph{Build Compliance Fact Constraints}

In addition to statutory constraints, each regulatory case provides factual conditions that must be formalized into logical predicates for reasoning. 
We construct a set of \textit{compliance fact constraints} $\mathcal{H}_{\text{case}}$ to represent the factual state of the case—such as capital adequacy ratios, remediation actions, or supervisory measures actually executed. 
These factual predicates are encoded in Boolean or numerical form and combined with the statutory constraint base $\mathcal{H}_{\text{law}}$ to form a complete logical representation of the case.

For the FSC case described above, the extracted factual inputs are:
\begin{tcolorbox}[colback=gray!5,colframe=gray!50!black,boxrule=0.3pt,
left=2pt,right=2pt,top=1pt,bottom=1pt]
\ttfamily\footnotesize
own\_capital = 111.09,\\
risk\_capital = 100.0,\\
net\_worth = 2.97,\\
plan\_submitted = true,\\
plan\_executed = false
\end{tcolorbox}
For variables not explicitly mentioned in the case description, the system treats them as \textit{free variables} rather than assigning default Boolean values.  
This design allows the solver to explore feasible valuations dynamically during constraint satisfaction, ensuring that the reasoning process remains faithful to the available factual scope while avoiding unwarranted assumptions.

These factual assignments constitute the case-specific constraint set $\mathcal{H}_{\text{case}}$, which, when merged with the law constraint base $\mathcal{H}_{\text{law}}$, produces the complete SMT problem instance:
\[
\mathcal{H}_{\text{law}} \cup \mathcal{H}_{\text{case}}.
\]

This unified representation allows the solver to verify whether the factual state of the case satisfies the statutory framework or violates one or more encoded provisions. 
It therefore provides a logically grounded test instance for subsequent consistency checking and compliance optimization.

\subsection{SMT Constraint Consistency Check and Optimization}
\subsubsection{Law Constraint Consistency Check.}
After encoding all statutory provisions as hard constraints, we perform a consistency validation to ensure the legal logic is self-coherent. The set of encoded legal formulas
\[
\mathcal{H}_{\text{law}}
\]
is submitted to the SMT solver. A satisfiable (\textit{SAT}) result indicates that the statutory provisions are logically consistent and contain no internal contradictions. This helps ensure that the encoded law forms a logically consistent foundation. Conversely, an \textit{UNSAT} result would reveal inconsistency within the legal encoding, prompting iterative refinement until the system achieves a consistent \textit{SAT} result.

\subsubsection{Case Illegality Check.}
Once the statutory constraints are verified to be internally consistent, we integrate the case-specific facts and perform a second verification phase. This Case Illegality Check determines whether the factual conditions correctly trigger a statutory violation by evaluating the combined constraint set:
\begin{equation}
\mathcal{H}_{\text{law}} \cup \mathcal{H}_{\text{case}} \cup \{\mathit{Penalty} = \mathrm{False}\}.
\label{eq:case-illegality}
\end{equation}
As shown in Eq.~(\ref{eq:case-illegality}), the combined constraint set incorporates both statutory and case-specific hypotheses, with the penalty condition explicitly fixed to \textit{False} to test for potential violations.

At this stage, an \textit{UNSAT} result is expected, indicating that the factual configuration is incompatible with lawful conditions and thus represents a valid non-compliant scenario. The solver demonstrates that no assignment can satisfy both the statutory and factual constraints simultaneously, confirming the existence of a legal violation.

For the FSC case, the solver indeed returns \textit{UNSAT}, revealing contradictions among:
\begin{tcolorbox}
\ttfamily\footnotesize
insurance:capital\_level,\\
meta:penalty\_conditions,\\
insurance:level\_3\_measures\_executed
\end{tcolorbox}
These entries form the \textit{unsat core}, pinpointing the minimal subset of constraints responsible for inconsistency. The conflict indicates violation of Articles~143-4 and~143-6 due to incomplete corrective measures, consistent with the actual administrative disposition.

If the solver instead returns \textit{SAT}, the logical formulation fails to capture the intended violation, and the system initiates an iterative refinement process to regenerate or adjust the encoded constraints until the expected \textit{UNSAT} outcome is achieved. The \textit{Case Illegality Check} therefore validates that the encoded case indeed represents a legitimate breach under the statutory framework, but it does not attempt to restore satisfiability or modify any factual predicates at this stage.

After a valid \textit{UNSAT} result is obtained, the system transitions to the optimization phase. In that subsequent step, Z3’s \textit{Optimize()} module is applied to the same constraint set, where case-specific predicates are relaxed into soft constraints while statutory provisions restored.

\subsubsection{Optimization for Minimal Compliance Solution}
Following the verification and adjustment phases described above, the optimization stage formalizes the search for the minimal compliance solution. 
Here, we treat the previously identified \textit{UNSAT} configurations as optimization targets and employ Z3’s \textit{Optimize()} module to derive the smallest set of factual changes that restore full logical satisfiability.

After constructing both hard and soft constraints, we employ Z3’s \textit{Optimize()} module to compute the \textit{minimal compliance solution}. This procedure searches for a model that satisfies all statutory (hard) constraints while modifying as few factual (soft) constraints as possible, thereby achieving legal compliance with minimal intervention. All factual statements extracted from a regulatory case are encoded as soft constraints and appended to the solver through the \code{add\_soft()} interface. Each soft constraint $s_i$ is assigned a weight $w_i = 1$, unless regulatory priorities specify otherwise (e.g., penalties may receive higher weights). The hard constraint set $\mathcal{H}$ represents immutable statutory obligations that must hold under all conditions.

The \textit{Optimize()} module operates internally through a weighted MaxSAT approach, aiming to maximize the total satisfaction of soft constraints while maintaining the logical consistency of all hard constraints. Formally, the optimization problem can be written as:
\begin{equation}
\max_{\mathbf{x}} \quad \sum_{s_i \in \mathcal{S}} w_i \cdot \mathrm{s_i}(\mathbf{x})
\quad \text{s.t.} \quad \mathcal{H}(\mathbf{x}) = \text{true},
\label{eq:maxsat}
\end{equation}
where $\mathrm{s_i}(\mathbf{x}) \in \{0,1\}$ indicates whether $s_i$ is satisfied under model $\mathbf{x}$. 

If no fully satisfiable assignment exists, Z3 incrementally relaxes violated soft constraints by introducing auxiliary Boolean selectors $\lambda_i \in \{0,1\}$  and minimizing their activation:
\begin{equation}
\min_{\mathbf{x}} \sum_{i=1}^{|\mathcal{S}|} w_i \lambda_i,
\quad \text{s.t.} \quad (\mathcal{H}(\mathbf{x}) \land \bigwedge_{i} (\mathrm{s_i}(\mathbf{x}) \lor \lambda_i))= \text{true}.
\label{eq:minlambda}
\end{equation}
As shown in Equation~\eqref{eq:maxsat}, the optimization seeks to maximize the satisfaction of all soft constraints under the statutory framework. When this is not achievable, Equation~\eqref{eq:minlambda} defines the relaxation procedure that minimizes the total violation cost by activating the smallest number of $\lambda_i$ variables. 

This mechanism effectively searches for the \emph{maximally satisfiable subset (MSS)} of all factual conditions, corresponding to the smallest logical modification required to restore legal compliance. Conceptually, Z3 iteratively relaxes the lowest-weighted constraints until all hard constraints are consistent, preserving maximal factual assumptions under the fixed statutory framework.

\vspace{3mm}
\begin{algorithm}[t]
\caption{Minimal Compliance via Weighted MaxSMT}
\label{alg:minimal_compliance_revised}
\begin{algorithmic}[1]
\Require Hard constraints $\mathcal{H}$; soft constraints $\mathcal{S} = \{s_i\}$ with weights $\{w_i\}$
\Ensure Minimal compliance model $\mathbf{x}^\ast$; violation set $\Delta$; cost $C(\Delta)$
\State $O \leftarrow \code{Optimize()}$ \Comment{Initialize MaxSMT optimizer}
\State Add all $h \in \mathcal{H}$ via $O.\code{add}(h)$
\State Add all $s_i \in \mathcal{S}$ via $O.\code{add\_soft}(s_i, w_i)$
\State Run $\text{status} \leftarrow O.\code{check()}$
\If{$\text{status} = \code{unsat}$}
    \State \textbf{return} No feasible compliance: hard constraints inconsistent
\EndIf
\State $\mathbf{x}^\ast \leftarrow O.\code{model()}$ \Comment{Optimal minimal-violation model}
\State Evaluate each $s_i$ under $\mathbf{x}^\ast$; collect $\Delta \leftarrow \{s_i \mid  {{s_i}(\mathbf{x}^\ast)=\text{false}\}}$
\State Compute $C(\Delta) \leftarrow \sum_{s_i \in \Delta} w_i$
\State \textbf{return} $\mathbf{x}^\ast, \Delta, C(\Delta)$
\end{algorithmic}
\end{algorithm}
\vspace{3mm}

In practice, the solver identifies minimal factual revisions—such as flipping a Boolean or adjusting a threshold—yielding $\mathbf{x}^\ast$ that satisfies $\mathcal{H}$ and minimizes $\sum_i w_i \lambda_i$. If $\mathcal{H}$ is inconsistent, the solver returns \code{unsat}.

For the FSC case, the optimization result suggests a single factual flip:
\begin{tcolorbox}[colback=gray!5,colframe=gray!50!black,boxrule=0.3pt,
left=2pt,right=2pt,top=1pt,bottom=1pt]
\ttfamily\footnotesize
improvement\_plan\_executed:\quad false $\rightarrow$ true
\end{tcolorbox}
This indicates that the firm would regain legal compliance by executing the improvement plan initially submitted, satisfying the requirements under Articles~143-4 and~143-6 of the \textit{Insurance Act}.

To enhance interpretability, the optimization results include both the optimal model $\mathbf{x}^\ast$ and the difference set $\Delta$, which lists the factual variables altered during optimization. Each element of $\Delta$ identifies which factual condition violated compliance and how it was adjusted, forming an explicit \textit{correction trace} interpretable by auditors or legal analysts. This trace directly supports post-hoc explanations and model accountability.

Each violated constraint indicates a specific non-compliant element, allowing transparent traceability and human-auditable reasoning. When integrated with the language model’s explanation module, each violation in $\Delta$ is translated into natural language statements such as: “The company failed to submit a remediation plan within the statutory period.” This hybrid reasoning procedure aligns with the legal principle of \textit{minimal intervention}, ensuring that compliance is restored through the smallest set of factual corrections while maintaining logical and statutory integrity.

The optimization aims to maintain logical soundness and interpretability, forming the foundation for transparent, verifiable compliance automation.

\noindent\textbf{Optimization Result}
When \textit{check()} returns \textit{sat}, the model $\mathbf{x}^\ast$ produced by \textit{Optimize()} minimizes $\sum_i w_i \lambda_i$ among all models satisfying $\mathcal{H}$. Therefore, the induced delta set $\Delta$ is a weighted-minimal set of factual violations. Under uniform weights, this coincides with a minimum-cardinality revision; with nonuniform weights, it yields a minimum-cost revision. In both cases, $\mathbf{x}^\ast$ constitutes the desired \textit{minimal compliance solution}.

\subsection{Multi-Agent Compliance Platform}
Building upon the constraint-generation and verification framework, we designed a modular multi-agent platform that operationalizes the reasoning process.
\begin{figure*}
    \centering
    \includegraphics[width=0.79\textwidth]{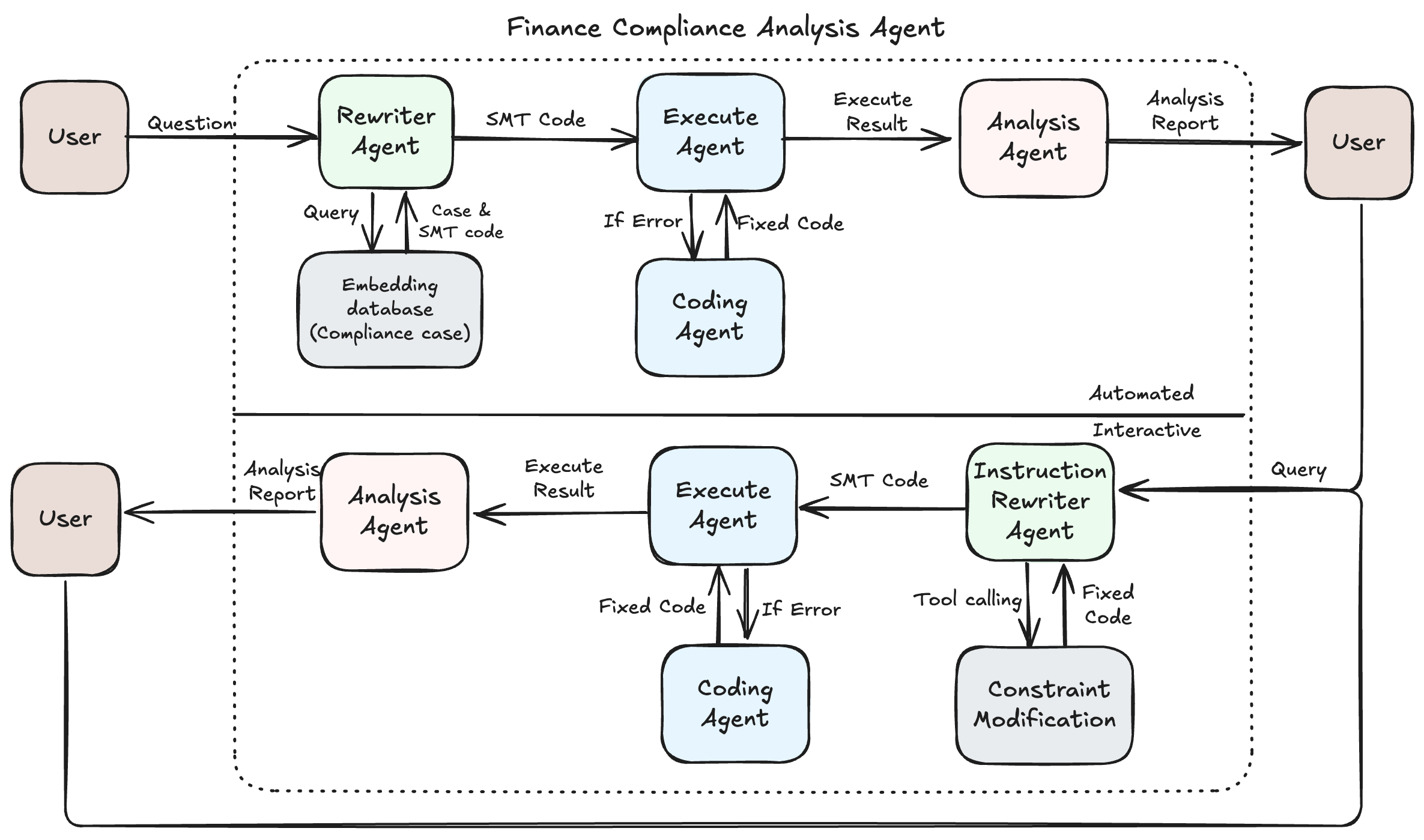}
    \caption{System architecture of the Finance Compliance Analysis Agent.}
    \label{fig:architecture}
\end{figure*}
\subsubsection{Design Principles}
The design of the Finance Compliance Analysis Agent adheres to four core principles: (1) Modularity, allowing agents to be replaced or upgraded independently; (2) Explainability, ensured through structured analysis and symbolic reasoning; (3) Human-in-the-loop adaptability, enabling iterative refinement via natural language; and (4) adaptability, supporting future expansion into additional regulatory domains or jurisdictions.
\subsubsection{System Architecture with AI Agents}

Figure~\ref{fig1} illustrates the architecture of the \textit{Finance Compliance Analysis Agent}, a modular multi-agent system designed to assess financial legal compliance. The workflow begins with a user query that sequentially passes through several specialized agents, each responsible for a distinct stage of reasoning and verification.

The Rewriter Agent reformulates the user query into a structured legal task and retrieves relevant precedents from the embedding-based \textit{Compliance Case Database}, which stores pre-generated SMT code for constraint evaluation. The Execute Agent then runs the retrieved SMT code using the SMT solver to verify compliance with the encoded legal constraints, supporting both automated evaluation and interactive refinement.

In cases where execution errors occur—such as syntax inconsistencies, ill-formed logic, or unbound variables—the Coding Agent collaborates with the Execute Agent to repair the SMT code. It inspects solver traces and conversation history to identify stable prior versions or generate corrective patches, thereby enabling a self-healing feedback loop between the solver and the agent.

The Analysis Agent interprets solver outputs and transforms them into structured compliance reports comprehensible to human auditors. When users provide feedback or suggest rule modifications, the Instruction Rewriter Agent translates these inputs into updated SMT logic for re-evaluation. Finally, the Modify Constraint Module performs low-level edits on SMT formulations, such as toggling logical conditions, injecting parameters, or fixing variable assignments.

Together, these coordinated agents form a robust and adaptable workflow that integrates symbolic reasoning, automated correction, and interactive human guidance for end-to-end legal compliance analysis.

\begin{table}[h]
\centering
\caption{Summary of LLM-generated SMT code}
\small
\begin{tabular}{lccc}
\toprule
\textbf{Method} & \textbf{Accuracy} & \textbf{Avg. Time (s)} \\
\midrule
LLM Generate SMT Code & 0.8621 & 36.85 \\
\bottomrule
\end{tabular}
\label{tab:llm_smt_generation}
\end{table}

\subsubsection{Agent Collaboration and Workflow}

The system operates through coordinated collaboration across reasoning stages. The Rewriter Agent reformulates user queries and retrieves relevant SMT cases, which the Execute Agent verifies via the SMT solver. When errors occur, the Coding Agent activates the self-healing mechanism—tracing the source of failure, performing rollback, or generating logical patches—until the constraints yield a valid satisfiable or unsatisfiable result.

Once successful execution is achieved, the Analysis Agent summarizes the results into interpretable compliance reports. For interactive exploration, users may modify assumptions or logic; the Instruction Rewriter Agent translates these inputs into structured tool calls dispatched to the Modify Constraint module, which updates and revalidates the SMT code.  

This modular, agent-oriented workflow ensures interpretability, fault tolerance, and adaptability, supporting both automated and human-in-the-loop compliance reasoning in evolving financial regulatory environments.

\section{Experimental Evaluation}

To validate our framework, we conducted an empirical analysis focusing on financial penalty cases issued by the FSC of Taiwan. 
In total, 87 real-world cases were collected and analyzed. 
We also developed an interactive user interface using Gradio to facilitate real-time interaction with the system. 
As part of our infrastructure, we curated a comprehensive regulatory database comprising 75 distinct laws and 3,753 statutory provisions extracted from the Taiwan Law Database.

This experiment is designed to address the following research questions:

\begin{itemize}
    \item \textbf{RQ1}: To what extent can LLMs effectively generate SMT constraints in the legal domain?
    \item \textbf{RQ2}: Does the integration of LLMs with SMTs enhance the model’s capability to reason about illegal terms in real-world regulatory cases, compared to using LLMs alone?
    \item \textbf{RQ3}: Does the integration of LLMs with SMTs improve the accuracy and consistency of legal compliance restoration in real-world cases, relative to standalone LLMs?
\end{itemize}

We evaluate these questions through both qualitative and quantitative assessments, focusing on reasoning performance, legal alignment, and correction accuracy.

All supplementary materials—including detailed SMT formulations, additional case studies, and implementation scripts— are publicly available on GitHub~\cite{nsc_github}. 

\begin{table*}[t]
\centering
\caption{Descriptive statistics of variables and constraints across 87 cases.}
\small
\begin{tabular}{lcccc|ccccc}
\toprule
 & \#Var& Real & Int & Bool  & \#Con & Hard & Soft  & H\_Ratio & S\_Ratio \\
\midrule
Average & 12.85 & 3.58 & 1.97 & 7.31  & 30.81& 19.30 & 11.51  & 0.63 & 0.37 \\
Min     & 4     & 0    & 0    & 1     & 11   & 6    & 4        & 0.38 & 0.24 \\
Max     & 40    & 21   & 7    & 21    & 63   & 39   & 25      & 0.76 & 0.63 \\
Std     & 5.60  & 3.40 & 1.74 & 4.11  & 9.62 & 6.35 & 4.61   & 0.08 & 0.08 \\
\bottomrule
\end{tabular}
\label{tab:case_stats}
\end{table*}

\subsection{RQ1: SMT Constraint Synthesis}

We conducted generation experiments on 87 regulatory cases. Using a few-shot learning approach, we prompted the LLM to generate SMT constraints tailored to each case. The generated code was then manually reviewed to verify its logical soundness and contextual correctness. If the code was valid, it was categorized as \textit{correct}; otherwise, we prompted the model to iteratively revise its output until the issues were resolved, and such cases were classified as \textit{incorrect}. Human inspection was conducted solely to verify factual consistency between the generated SMT constraints and the original case context. No manual editing or correction was performed on model outputs, and to ensure fair comparison, human intervention was limited to factual verification only—without influencing any intermediate reasoning or solver correction steps. This procedure guarantees that all evaluations objectively reflect the model’s intrinsic capability.

Common errors included omissions of impacted clients in compliance outcomes, failure to zero out fines where applicable, and logically inconsistent constraint sets resulting responses from the solver. Through manual correction and validation, we ensured that all final SMT codes were accurate and could serve as ground truth for downstream evaluation.

The generation accuracy and response times are summarized in Table~\ref{tab:llm_smt_generation}.

\textbf{Scale Analysis of Case Studies.}
To clarify the scale of our case studies, we analyzed the number of variables and constraints across all 87 cases. 
The detailed statistics are summarized in Table~\ref{tab:case_stats}. On average, each case contains about 13 variables (\#Var) and 31 constraints (\#Con), with the largest instance involving 40 variables and 63 constraints. Roughly 63\% of the constraints are hard constraints (H\_Ratio), while 37\% are soft (S\_Ratio), reflecting a balance between strict regulatory rules and flexible planning considerations.

\begin{figure}[htbp]
  \centering
  \subfloat[Variables vs. Constraints]{%
    \includegraphics[width=0.48\linewidth]{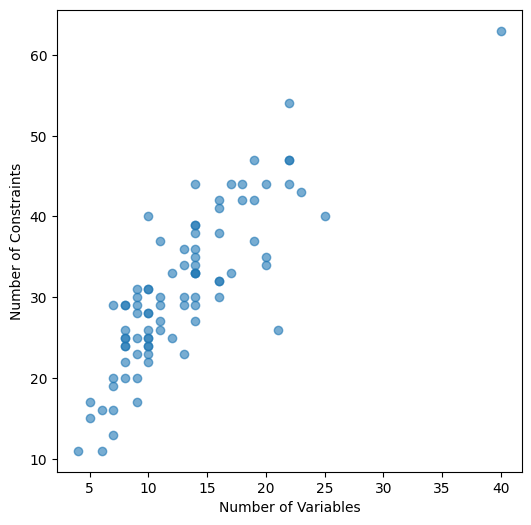}%
    \label{fig:sub_vars_constraints}
  }
  \hfill
  \subfloat[Constraint Distribution]{%
    \includegraphics[width=0.48\linewidth]{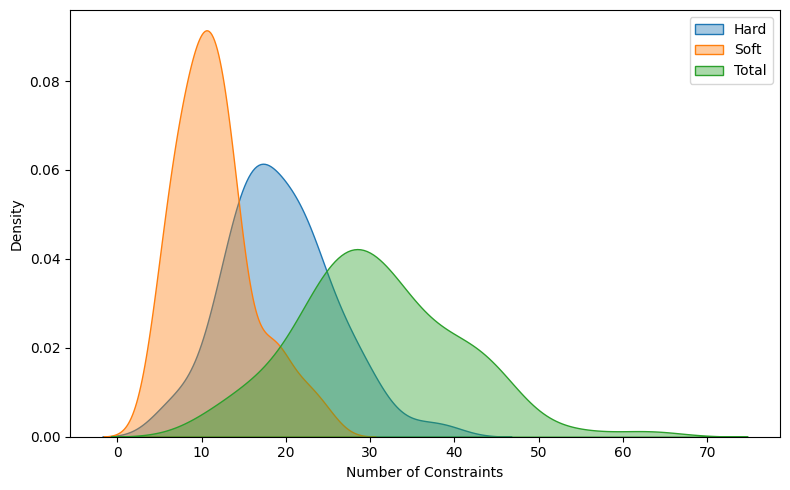}%
    \label{fig:sub_constraint_dist}
  }
  \caption{Relationship between variables and constraints (a), and the distribution of hard, soft, and total constraints (b).}
  \label{fig:combined_constraints}
\end{figure}

As illustrated in Fig.~\ref{fig:combined_constraints}, the scatter plot between the number of variables and constraints reveals a clear positive correlation. This observation indicates that as the number of variables increases, the constraint set grows proportionally, suggesting that the proposed framework is suitable for small- to moderate-scale SMT problems.

Fig.~\ref{fig:combined_constraints} further shows the distribution of hard, soft, and total constraints. The hard constraints are primarily concentrated between 15 and 25, while the soft constraints fall mostly within the range of 8 to 15. Consequently, the total number of constraints typically ranges from 25 to 40, with a few outliers exceeding 60.

\begin{table}[h]
\centering
\caption{Illegal term analysis: Comparison of LLM and LLM+SMT.}
\small
\begin{tabular}{lcc}
\toprule
\textbf{Method} & \textbf{Avg. Illegal Terms per Case} & \textbf{Avg. Time (s)} \\
\midrule
LLM + SMT & 5.08 & 0.021 \\
LLM Only & 6.40 & 7.01 \\
\bottomrule
\end{tabular}
\label{tab:illegal_term_analysis}
\end{table}

\begin{table*}[t]
\centering
\caption{Performance of case Compliance Restoration.}
\small
\begin{tabular}{lccccc}
\toprule
\textbf{Method} & \textbf{Accuracy} & \textbf{Precision} & \textbf{0} & \textbf{F1 Score} & \textbf{Avg. Time (s)} \\
\midrule
LLM + SMT & 1.0000 & 1.0000 & 1.0000 & 1.0000 & 0.0040 \\
LLM Only & 0.3333 & 0.6190 & 0.2063 & 0.3080 & 1.4700 \\
\bottomrule
\end{tabular}
\label{tab:correction_results}
\end{table*}


\subsection{RQ2: Illegal Term Reasoning}

To support compliance in real-world legal cases, our system is designed to identify specific illegal terms—statutory violations—within each scenario. Highlighting these terms helps users recognize which actions must be taken to align with relevant regulations. We evaluated this capability by comparing the number of identified illegal terms and the average response time between the baseline LLM and our proposed LLM+SMT approach, as shown in Table~\ref{tab:illegal_term_analysis}.

Our SMT-based method leverages \textit{assert} and \textit{track} commands during code generation. When contradictions arise—i.e., Z3 returns an \textit{unsat} result—this indicates a violation between statutory requirements and the case description. For instance, if the law mandates remedial actions for capital inadequacy but the case omits such steps, the system flags this inconsistency.

\begin{table}[h]
\centering
\caption{Confusion matrix for LLM without SMT Solver.}
\small
\begin{tabular}{lcc}
\toprule
\textbf{Prediction} & \textbf{Actually Satisfiable} & \textbf{Actually Unsatisfiable} \\
\midrule
Satisfiable          & 13 (TP) & 8 (FP) \\
Unsatisfiable        & 50 (FN) & 16 (TN) \\
\bottomrule
\end{tabular}
\label{tab:confusion_matrix}
\end{table}

Since Z3 returns only a minimal unsat core by default, we redefine all constraints—including originally soft ones—as hard constraints, forcing the solver to expose multiple unsat cores. We then extract all components traced to hard constraints across iterations to define the full set of "illegal terms." Remaining elements from soft constraints are treated as free variables and iteratively reintroduced until the system reaches a satisfiable state. The total number of unique illegal terms is computed via set union over these rounds.

To validate this mechanism, we also queried the LLM directly using the same case and legal text. However, the LLM's output tends to lack logical consistency and interpretability. By contrast, our SMT-enhanced method not only produces comparable results but does so with over two orders of magnitude improvement in reasoning efficiency and improved explainability.

In a real insurance case,\footnote{See the FSC penalty announcement for Sang Shang Meibang Life Insurance Co., Ltd.: \url{https://www.fsc.gov.tw/ch/home.jsp?id=131&parentpath=0,2&mcustomize=multimessages_view.jsp&dataserno=202407120004&dtable=Penalty}}  our SMT module identified the following illegal terms in 0.191 seconds: \code{req\_net\_worth}, \code{self\_capital\_sum}, \code{risk\_capital\_sum}, \code{def\_improvement\_plan}, \code{req\_final\_CAR}, \code{def\_pre\_CAR}, and \code{def\_total\_improvement}. These correspond to formal statutory violations that must be resolved to restore compliance.

In contrast, the LLM took approximately 5 seconds and returned a longer, less precise list, including: Capital Adequacy Ratio below statutory level, Net Worth Ratio below regulatory requirement, Significant capital inadequacy, Failure to submit a feasible capital improvement plan, Failure to ensure capital adequacy in 2024, High uncertainty in planned private placement of hybrid instruments, Inadequate backup plan, Violation of Insurance Act Article 143-6 Paragraph 1 Subparagraph 1, Restricted transactions with related parties, and Noncompliance with capital adequacy regulations.

Closer inspection shows these additional terms fall into three categories: (1) \textbf{Contextual Misclassifications}—observations that indicate financial risk but are not legal violations (e.g., uncertainty in private placement); (2) \textbf{Misinterpretation of Advisory Clauses}—soft recommendations wrongly flagged as mandatory (e.g., lacking a contingency plan); and (3) \textbf{Vague or Composite Labels}—overgeneralized statements that lack traceability to specific provisions.

These findings highlight the key strengths of our SMT-based approach: precise legal modeling, symbolic consistency checking, and the ability to isolate actionable violations. In domains such as financial regulation, where auditability and precision are critical, this advantage is particularly valuable.
\subsection{RQ3: Compliance Restoration}

Following the analysis of illegal term identification, we proceed to evaluate the models’ capabilities in correcting real-world cases. This task investigates what changes are necessary to revise a case such that the resulting legal outcome would be penalty-free. We performed this correction analysis on all 87 cases.

For the SMT-based approach, we increased the weight of the penalty-related soft constraints and set the target penalty value to zero. This allows the SMT solver to return a satisfiable assignment that eliminates the penalty, which serves as the \textit{ground truth} for comparison. Importantly, the ground truth in this task does not refer to a single unique assignment but rather to the satisfiability status (\textit{SAT} or \textit{UNSAT}) of the case. For satisfiable instances, multiple valid assignments may exist, and any assignment that satisfies all constraints is accepted as correct. Thus, our evaluation does not require the LLM to replicate a specific SMT-generated assignment; it only needs to produce a valid satisfiable solution when one exists.

In total, 63 cases were satisfiable, among which only 13 were accompanied by a sample assignment for reference, while the others were not restricted to a single solution. This setup aligns with real-world legal reasoning, where multiple compliant revisions of a case may coexist.

In this experiment, the LLM is provided with the soft constraint variables and their original values. It is allowed to freely decide whether to flip the value of a given variable or assign a new one. The LLM also receives the corresponding case text and statutory references to support its reasoning process.

To further evaluate model robustness, we randomly selected 24 cases and introduced contradictory constraints into the solver. For example, we may define a capital adequacy ratio requirement of over 200\%, where the ratio is calculated as own capital divided by risk-weighted capital. If we then add a constraint setting own capital to zero, the system becomes unsatisfiable. These newly introduced constraints are also provided to the LLM. If the LLM correctly reasons that the problem has no feasible solution, it should return an unsatisfiable judgment.

The performance comparison is shown in Table~\ref{tab:correction_results}, and the confusion matrix for the LLM-only baseline is shown in Table~\ref{tab:confusion_matrix}.

From the results, we observe that the LLM tends to default to predicting an unsatisfiable outcome, even for cases where a valid solution exists. This bias leads to a significant number of false negatives. Furthermore, the LLM’s compliance restoration lacks consistency and completeness compared to the SMT solver. In contrast, the SMT-based method achieves high accuracy in compliance correction tasks and substantially faster inference, completing each correction in approximately 0.004 seconds—over 300 times faster than the average LLM response time.

\begin{table*}[!b]
\centering
\caption{Internationalization Potential: Minimal Engineering Modifications Required for Legal Domain Transfer}
\label{tab:internationalization_costs}
\small
\begin{tabular}{llc}
\toprule
\textbf{Module} & \textbf{Purpose} & \textbf{Should it be modified?} \\
\midrule
RAG Corpus & Retrieve relevant legal text and precedents & Yes (It should be replaced with a local legal corpus.) \\
Prompt Template & Control query format and language context & Yes (It should be adjusted for language and legal phrasing.) \\
Clause Parser & Map legal clauses to logical templates & Partial (Parsing patterns should be redefined.) \\
Legal Mapping Dictionary & Link clause labels to statutes & Yes (It should be updated for the target jurisdiction.) \\
Z3 Rule Engine & Enforce symbolic legal constraints & No (It can be reused without modification.) \\
SMT Relaxation Loop & Perform minimal correction search & No (It remains domain-independent.) \\
Gradio UI & Provide user interface & No (It can be directly reused.) \\
\bottomrule
\end{tabular}
\end{table*}

\subsection{Summary of Experimental Findings}
Across all three research questions, the results consistently demonstrate the effectiveness of integrating SMT solvers into the legal reasoning pipeline. For \textbf{RQ1}, the LLM generated syntactically and semantically correct SMT code in over 86\% of cases, enabling precise modeling of legal constraints. In \textbf{RQ2}, the SMT-enhanced method achieved comparable illegal term identification to the baseline LLM while offering faster responses and greater logical clarity. For \textbf{RQ3}, the Compliance Restoration powered by LLM+SMT reached perfect accuracy, outperforming the baseline across all major metrics. Overall, the hybrid LLM+SMT framework proves both robust and interpretable, showing strong potential for practical legal reasoning and compliance correction.

\vspace{6pt}
\textbf{Cross-Jurisdictional Adaptability.} Although our evaluation focuses on Taiwan’s financial regulatory environment, the proposed framework is \textit{jurisdiction-agnostic}. By adapting the retrieval corpus and prompt exemplars to reflect the case precedents and statutory structures of target jurisdictions, the system can be extended to common law settings such as the United States or the United Kingdom. Future work will explore multilingual and precedent-driven adaptations to further evaluate cross-jurisdictional applicability.

\section{Conclusion}

We propose \textbf{Neuro-Symbolic Compliance}, a hybrid framework integrating LLM-based understanding with SMT-based reasoning through a multi-agent design. Each agent models domain-specific expertise, enabling interpretable and verifiable analysis of financial regulation. Applied to Taiwan’s FCS cases, our system outperforms LLM-only baselines in reasoning consistency and inference speed. This work demonstrates how agent-based neuro-symbolic AI can bridge natural legal language and machine-verifiable logic toward scalable regulatory automation.

\section*{Acknowledgement}
This work is partially supported by the National Science and Technology Council (NSTC), Taiwan, under Grant No. 113-2221-E-004-010-MY2 and 114-2221-E-002-183-MY3, and by the NTU-Delta Electronics Innovation Research Funding Project under Grant No. NTU-DE-114FR032, and by Chunghwa Telecom Co., Ltd. under the commissioned project “National Chengchi University Financial AI Research Project: Application of Traditional Chinese Large Language Models and Artificial Intelligence Agent Platforms for Investment Analysis Report Generation.”

\bibliographystyle{IEEEtran}
\bibliography{reference}
\vspace{12pt}
\end{document}